\documentclass[nonacm]{acmart}

\setcopyright{none}
 \renewcommand\footnotetextcopyrightpermission[1]{}

\usepackage[caption=false,font=footnotesize]{subfig}
\usepackage{multirow}
\usepackage{algorithm}
\usepackage{algpseudocode}
\usepackage{amsmath}
\AtBeginDocument{%
  }

\makeatletter
\let\@authorsaddresses\@empty
\makeatother

\begin{document}

\title{FCx: An algorithm for finding Feasible Counterfactual Explanations}

\author{Markou Kleopatra}
\email{klmark@di.uoa.gr}

\affiliation{%
  \institution{National and Kapodistrian University of Athens,}
  \institution{Department of Informatics and Telecommunications}
  \city{Athens}
  \country{Greece}
}

\author{Vana Kalogeraki}
\affiliation{%
  \institution{Athens University of Economics and Business,}
\institution{Department of Informatics}
  \city{Athens}
  \country{Greece}}
\email{vana@aueb.gr}

\author{Gunopulos Dimitrios}
\affiliation{%
  \institution{National and Kapodistrian University of Athens,}
  \institution{Department of Informatics and Telecommunications}
  \city{Athens}
  \country{Greece}
  \email{dg@di.uoa.gr}
}

\begin{abstract}
Counterfactual (CF) explanations identify changes that alter an input’s classification. While existing methods produce realistic and low-cost CFs, they often fail to ensure feasibility, by suggesting non-constructive modifications or incompatible with future changes (e.g., changing an individual's race to secure a job offer). We introduce a refinement of CF explanations that explicitly enforces feasibility. Our approach is the first to efficiently generate CFs that are realistic, low-cost and feasible. We accommodate both hard feasible constraints, specified by domain knowledge users, and soft feasible constraints, inferred automatically via causal inference from the dataset. Our method, \textit{Feasible Counterfactual Explanations (FCx)}, is based on a modified Variational Autoencoder (VAE) optimized with a multi-factor loss function. We measure the cost of a change based on the absolute change in values (proximity) as well as the number of features changed (sparsity) while realism is measured based on the LOF for density estimation, guaranteeing that CFs reside in densely populated regions. Extensive experiments on four public datasets show that our approach matches state-of-the-art performance across multiple metrics while guaranteeing feasibility.
\end{abstract}


\maketitle

\section{Introduction}
Understanding predicted output is a crucial element of any Machine Learning (ML) model.
Researchers often question whether a model’s predictions are interpretable and applicable to real-world scenarios. In various fields, including criminal justice \cite{criminaljustice}, clinical healthcare \cite{CRESSWELL20173}, loan approvals, and hiring decisions, the ability of a system to provide a valid explanation is essential for guiding future actions. In addition, \cite{Wachter} displays in detail the legal rights that an individual has, regarding an explanation over the output, since its personal data has been used in similar occasions. 
Given these considerations, it is crucial to make ML models more interpretable and provide individuals with actionable insights on how to achieve a desired prediction outcome \cite{Miller}. Counterfactual (CF) explanations \cite{Wachter} serve as a widely used form of local explanations, as they align with the ML model while remaining interpretable. A local CF explanation refers to minor perturbations in the input features that remain close to the original instance but result in a different model prediction. 

\begin{figure}[!ht]
    \centering
    \includegraphics[width=0.67\textwidth]{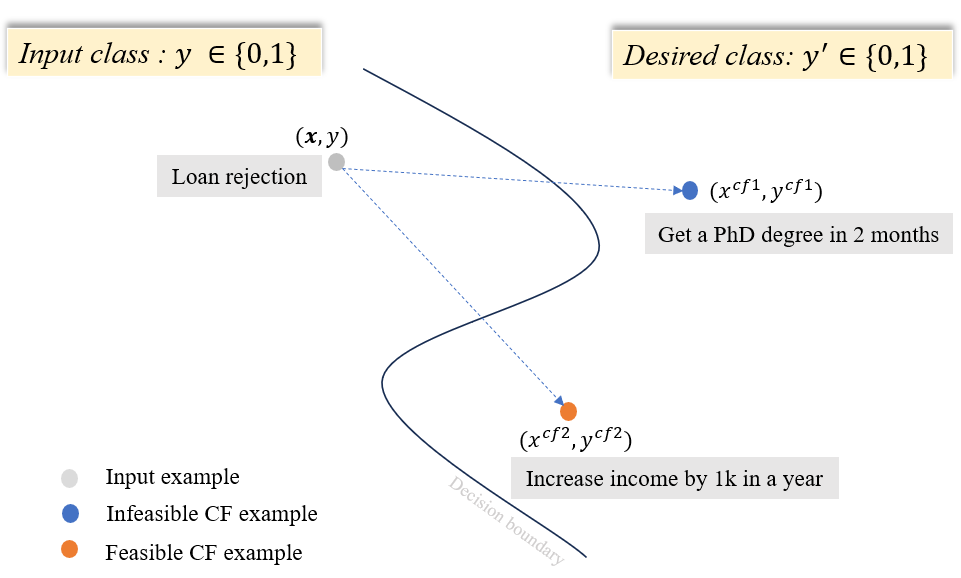}
    \caption{Illustrative example of possible Counterfactual Explanations feasible and infeasible}
    \label{fig:cfexampledefinition}
\end{figure}

A commonly used example to explain this concept answers the question, "What changes should an individual make to qualify for a loan that is currently denied?" Any potential modification that shifts the model’s decision from the original class to the desired one is considered a CF example. Figure \ref{fig:cfexampledefinition} presents a visual illustration of two counterfactual examples, both of which are technically valid; however, their classification as feasible or infeasible is determined by the extent to which they satisfy the predefined constraints of the model.

However, there are several questions arisen regarding these examples. Are actually all these modifications applicable to the real world? Are all CF explanations easily adaptable to real-world applications? The answer is negative to these questions. A Machine Learning model can learn from data distributions to recognize patterns and try to imitate them, but it lacks logic. Its produced results cannot apply to a multifactorial real-world scenario. Therefore, it is quite common to see CF examples suggesting unachievable goals (see Figure \ref{fig:cfexampledefinition}). For example, a hiring application would propose to a candidate to obtain a degree in a two-month period or extreme cases become younger to get a particular job, while both suggestions are infeasible and useless to the user. Resolving feasibility is essential because counterfactual explanations are often interpreted as practical guidance on how to achieve a desired outcome (e.g., securing future loan approval). When counterfactuals are unattainable or depend on model behavior in sparse, unreliable regions of the feature space, they can mislead and in some cases even offend those who receive them, undermining trust and reducing their real-world usefulness\cite{FACE,DiCE,vermaREVIEW}.

We approach \textit{feasibility} by enforcing constraints that promote (i) real-world applicability, (ii) realism, and (iii) consistency with causal constraints among features, thereby yielding counterfactual scenarios that are both valid and meaningful. There are two primary approaches to formulating such constraints. First, domain experts can provide prior knowledge to establish logical constraints manually \cite{Mahajan} (hard causal feasibility constraints). Second, causal discovery algorithms can be used to identify existing dependencies within a dataset, enabling the construction of a full or partial causal graph (soft causal feasibility constraints). In this work, we introduce a straightforward methodology that effectively generates CF examples while ensuring compliance with both types of causal constraints.

\begin{figure}[!ht]
    \centering    \includegraphics[width=0.67\textwidth]{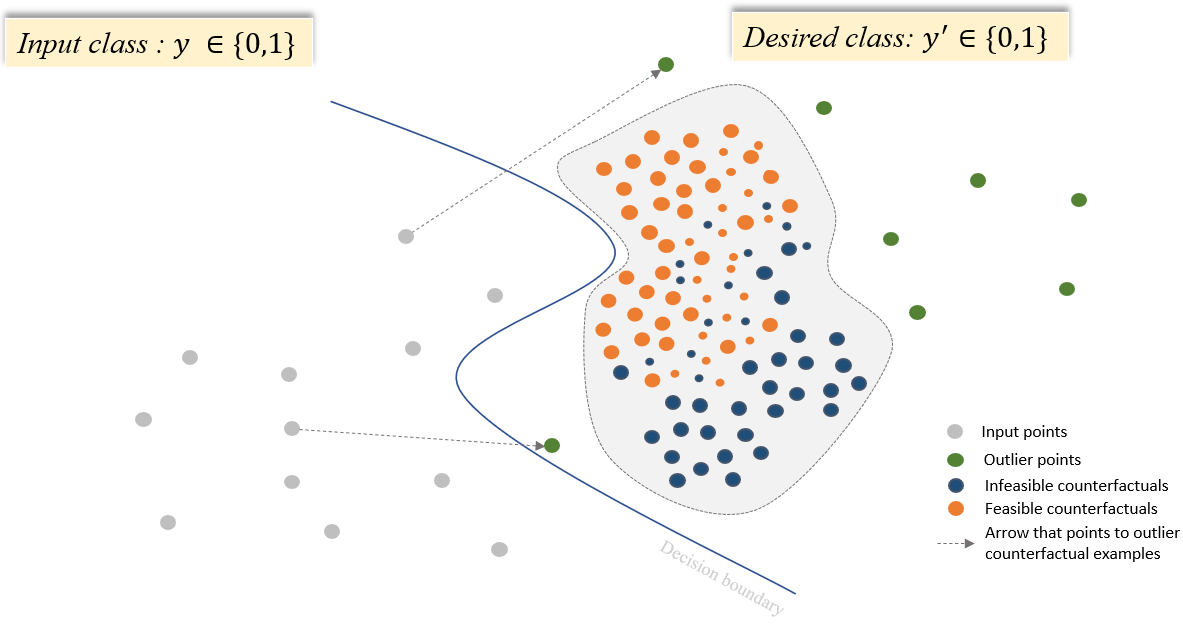}
    \caption{Illustrative example that depicts the existence of outlier CF examples outside densely populated regions}
    \label{fig:cfexampleanddensity}
\end{figure}

A CF example may be potentially feasible but unlikable based on the probability distribution that we have seen. It is important to provide CF examples that are highly probable to reside in densely populated regions.  
So, to make our model robust, in addition to feasibility, we employed other criteria that can be used to determine whether we are going to select a CF example or not. In our work, we use a statistical state-of-the-art technique that helps us first estimate the density of the latent space and then learn to generate CF examples in these dense regions eliminating outliers.
Figure \ref{fig:cfexampleanddensity} demonstrates one of our goals, the identification of dense regions where it is more likely a point to be a feasible CF example. Our network is trained to generate CFs examples in densely populated regions (grey area) and to exclude the outliers. The \textit{orange} dots represent the feasible CF examples, the \textit{blue} dots represent the infeasible CF examples and the \textit{green} dots are the outliers, respectively.  
This technique can strengthen the probability of finding feasible CF examples when they learn to be located in much more dense regions, rather than being spread in the entire space. We chose to use the density-based LOF technique \cite{LOF_Breunig2000} to eliminate CF outliers. We apply the technique in the lower dimensionality latent space $z$ of a Variational Autoencoder model \cite{autoencoding} because density based techniques work better in lower dimensionality spaces. Intuitively, our goal is to produce CF examples that are likely to be real examples and be located in dense regions with examples. This is achieved after training the VAE and formulating its loss function to satisfy our desiderata. 

Counterfactual (CF) explanations must offer individuals \textit{actionable} alternatives, helping them navigate toward solutions tailored to real-world occasions. However, an important consideration is determining the point at which a solution becomes impractical due to excessive modifications. Users typically prefer achieving their desired outcome with minimal adjustments. This highlights the need for a metric that quantifies the number of changes required.

\begin{figure}[!ht]
    \centering
    \includegraphics[width=0.67\textwidth]{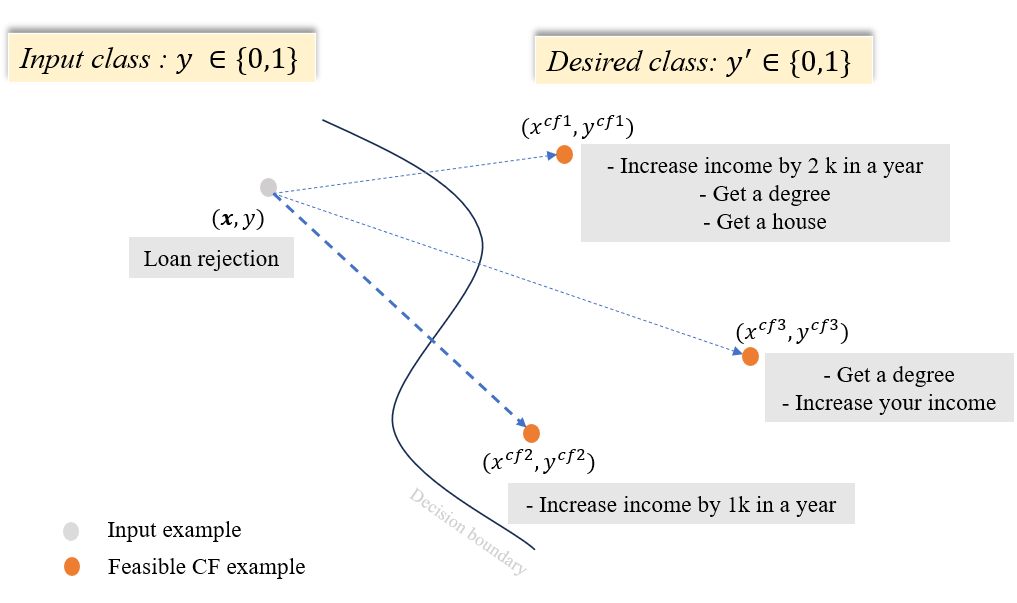}
    \caption{Illustrative example of possible reconstructed feasible suggestions with a preference (bold dashed arrow) to the one with the least amount of changes for the end-user}
    \label{fig:cfexampleandsparsity}
\end{figure}

To address this, we incorporate a sparsity constraint into our methodology, improving the model’s robustness while reducing unnecessary alterations \cite{Sparsity}. As illustrated in Figure \ref{fig:cfexampleandsparsity}, three feasible CF examples suggest different ways an individual could qualify for a loan. The most desirable CF example is the one that requires the fewest modifications (represented by the orange dot, highlighted with a blue dashed line). In this work, we introduce sparsity as a guiding principle, enabling the model to generate CF explanations that minimize the cost of achieving the desired outcome.

Finally, in this paper, our main contributions can be summarized as follows: 
\begin{itemize}

\itemsep0em 
    \item  We introduce Feasible Counterfactual Explanations (\textbf{FCx}), a novel approach to generate counterfactual efficiently (CF) explanations that ensure realism, low-cost modifications and feasibility, incorporating both hard feasibility constraints (from domain knowledge) and soft feasibility constraints (inferred via causal inference).
    \item We incorporate sparsity into our model, aiming for low-cost modifications through the fewest possible changes when generating CF examples.
    \item Additionally, we ensured that generated CFs reside in densely populated regions, avoiding unrealistic or outlier cases. To do that, we used the Local Outlier Factor (LOF) technique in reduced dimensionality space.
    \item FCx manages to effectively integrate diverse elements, such as causal constraints, sparsity, and density-based validation, to adapt to varying dataset characteristics and ensure the generation of realistic and feasible counterfactual examples.
\end{itemize}

\section{Related work}
\label{sub:related}

The research surrounding explainable machine learning (XAI) has grown vastly these past years\cite{bodriaXAI,vermaREVIEW,karimi2021survey,CEM,DiCE,REVISE,C-CHVAE,Mahajan, Papapetrou1,Salimi_2023}, with counterfactual (CF) examples receiving significant attention. \cite{vermaREVIEW} provides a comprehensive review of the algorithmic diversity in this field, identifying existing gaps and outlining future research directions. Additionally, \cite{karimi2021survey} presents a collection of methods focused on algorithmic recourse, which aims to systematically reverse unfavorable algorithmic decisions by offering explanations and actionable recommendations for affected individuals.

Some approaches move beyond traditional supervised learning techniques and incorporate self-supervised methods. For instance, \cite{CEM} employs contrastive learning to justify black-box classifier predictions. Other studies explore different aspects of CF explanations: \cite{DiCE} emphasizes on the diversity of CF examples and has been implemented as a Python library for generating them, while \cite{REVISE} focuses on producing actionable CF examples by minimizing the number of required changes. The work in \cite{C-CHVAE} highlights the importance of faithfulness in CF explanations by balancing proximity (ensuring CF examples are not local outliers) and connectedness (ensuring CFs remain close to valid input values). Similarly, \cite{Mahajan} prioritizes feasibility by leveraging structural causal models. 

Further contributions include \cite{rugolon_papapetrou}, which evaluates CF explanations based on sparsity and LOF. Sparsity measures the minimal changes required for a patient to achieve a positive outcome, while LOF assesses whether the generated CF remains close to the data distribution of the desired class. Feasibility on synthetic and image datasets is investigated by \cite{FACE}, which leverages high-density paths within the data manifold to propose achievable and realistic counterfactual changes. Additionally, feasibility is studied at both the population and individual levels: \cite{FACEGROUP} generates feasible group/subgroup counterfactuals for fairness auditing, while \cite{BOVE} examine how to present multiple feasible counterfactual options for a single instance.

More recent works have further explored generative and causality-aware mechanisms for producing realistic and feasible CF explanations. In \cite{FCEGAN}, the authors introduce FCEGAN, a generative framework based on counterfactual templates that allows users to specify mutable features at inference time, enabling flexible CF generation under user-defined constraints. Similarly, \cite{VCNet} proposes VCNet and ImmutableVCNet to generate realistic class-consistent CFs using conditional variational autoencoders, while \cite{CEILS} formulates CF explanations as interventions in latent space in order to incorporate causal relations and provide feasible recommendations.

\section{Problem Definition}
\label{prob_definition}

In this section, we provide an in depth presentation of our problem definition, followed by the basic components of our model. 

Let $\mathbf{x} \in \mathbb{R}^d$, $y \in\{0,1\}$ represent the input data point with $d$ features and $y$ the input class. 
Given an example $(\mathbf{x})$ and a target class $y\prime$ we would like to generate a CF example to minimize the following:

\begin{equation}
\label{definition}
\begin{aligned}
\operatorname{min}_{x^{cf}\in\mathcal{C}} \operatorname{Dist}(x,x^{cf}) + \sum_{(p, q) \in \mathcal{C}} \text {Penalty\_h}\left(x^{cf}_p, x^{cf}_q\right) + \\
+\sum_{(p, q) \in \mathcal{C}} \text {Penalty\_s}\left(x^{cf}_p, x^{cf}_q\right) + \\
+ \left\|\mathbf{x}^{cf}-\mathbf{x}\right\|_0 + \\
+ \text{LOF}(x^{cf})    
\end{aligned}
\end{equation}

such that $h(x^{cf}) = y\prime$,
where $h$ refers to a trained classifier and $h(x^{cf})$ indicates the predicted class of the cf example.

Breaking down the components of the Equation \ref{definition}, we have: 
\begin{itemize}
    \item $\text Dist$ that refers to a distance measurement such as $L1/L2$ distance
    \item $\mathcal{C}$ is a set of causal relationships between pairs of features with $(p,q)\in\mathcal{C}$ represents a directed relationship between $x_p \longrightarrow x_q$
    \item $x^{cf}_p$ and $x^{cf}_q$ are the cf values of the features $p$ and $q$, respectively, 
    \item $\text{Penalty\_h}(x^{cf}_p, x^{cf}_q)$ that measures how much the values $x^{cf}_p$ and $x^{cf}_q$ violate the hard causal relationship between them
    \item given an adjacency matrix $A$ extracted from a causal graph which depicts the relationships between features in a dataset, there is a $\text{Penalty\_s}(x^{cf}_p, x^{cf}_q)$ which encourages zero covariance between non-connected pairs of features and non-zero covariance for connected ones, to measure every soft causal relationship between features
    \item $\left\|\mathbf{x}^{cf}-\mathbf{x}\right\|_0$ which penalizes the number of feature changes
    \item $\text{LOF}(x^{cf})$ which encourages $x^{cf}$ to be in a dense region of a data distribution where LOF indicating how isolated $x^{cf}$ is from high-density regions. 
\end{itemize} 

\subsection{Feasibility and Density estimation}
\label{Feas_lof}

Feasibility is the most critical aspect of our model, with various components from the previous problem definition (see Section \ref{prob_definition}) contributing to its formulation. We consider a CF example feasible when the following definition is satisfied. 
 
\begin{definition} \textit{Feasibility}: Let the ($\mathbf{x}$, $y$) be the input features and the predicted outcome of the model and $y’$ be the desired output class. We can define a CF example ($x^{cf}$, $y^{cf}$) as \textit{feasible}, if the desired class $y’$ is equal to the $y^{cf}$ output, the changes from $x$ to $x’$, satisfies all constraints provided by domain knowledge and all variables that conduct a \textit{causal model} (i.e. a structure that depicts all possible relations between the variables of a dataset), lie within the input domain \cite{Mahajan}.
\end{definition}

What if a CF example can be characterized as feasible according to the previous Definition 3.1, following every condition that is proposed (distance, validity and the hard causal constraints), but is located far from other CF points in the example space? Certainly, there will be such points. Our model tries to eliminate such points considering them as outliers. That's because these points are located outside the identified dense regions and consequently the probability of them being feasible is more likely to be significantly smaller.  

To efficiently eliminate such outlier CF examples and identify more densely populated regions, we use a state-of-the-art density-based outlier detection technique, specifically the statistical LOF and we used it both as a learnable parameter and as a metric. For completeness, we include the definition of LOF provided by \cite{LOF_Breunig2000}.

\begin{definition}\textit{Local Outlier Factor}: is a density-based technique that uses the nearest neighbor search to identify the anomalous points \cite{LOF_Belyadi}. The advantage of using LOF is identifying points that are outliers relative to a local cluster of points. For instance, when using the LOF technique, neighbors of a certain point are identified and compared against the density of the neighboring points. So, given a set of points in the counterfactual space $ X^{cf}=\left\{x^{cf1}, x^{cf2}, \ldots, x^{cfn}\right\}$, we compute the pairwise Euclidean distances between all points in the set and then apply the KNN method to find the k-nearest neighbors for each point. Based on this KNN distances we calculate the \textit{Reachability distance} (\textit{rd}), as denoted 
in Equation \ref{reachability distance} below:

\begin{equation}
\label{reachability distance}
{\operatorname{rd}_k\left(x^{cfi}, x^{cfj}\right)}=\max \left(\operatorname{dist}\left(x^{cfi}, x^{cfj}\right), \mathrm{k} \text {-dist}\left(x^{cfj}\right)\right)
\end{equation}

where $\mathrm{k} \text {-dist}(x^{cfj})$ is the distance from $x^{cfj}$ to the k-\textit{th} nearest neighbor. So, the \textit{Local Reachability Density (LRD)} is formed as follows in Equation \ref{lrd} and computes the local reachability density for each point. 

\begin{equation}
\label{lrd}
\operatorname{LRD}_k\left(x^{cfi}\right)=\left(\frac{\sum_{x^{cfj} \operatorname{knn}\in\left(x^{cfi}\right)} \operatorname{rd}_k\left(x^{cfi}, x^{cfj}\right)}{k}\right)^{-1}
\end{equation}

Given the LRD, the LOF is formed as denoted in Equation \ref{lof} and is defined as the average ratio of the local reachability density of $x^{cfi}$ to those of its k-nearest neighbors. 

\begin{equation}
\label{lof}
\operatorname{LOF}_k\left(x^{cfi}\right)=\frac{\sum_{x^{cfj} \in \operatorname{knn}\left(x^{cfi}\right)} \frac{\operatorname{LRD}_k\left(x^{cfj}\right)}{\operatorname{LRD}_k\left(x^{cfi}\right)}}{k}
\end{equation}

Finally, our goal is to achieve LOF $\approx 1$, which can be translated into dense regions with normal points, while LOF $\gg 1$ indicates the existence of outliers.
\end{definition}

\subsection{Hard Causal constraints}
\label{causalconstraints}

Suppose we establish constraints to define the feasibility of a counterfactual (CF) example by leveraging a causal model or fundamental domain knowledge specific to a dataset and its attributes. For instance, in the context of loan applications, if individuals must adjust certain factors to qualify, a CF example that reduces the "age" attribute would be characterized as unrealistic. This is because it contradicts the causal rule that age can only increase over time \cite{Mahajan}.

To make our approach adaptable to various objectives, we rely on logical constraints derived from basic domain knowledge. To establish a foundational framework, we first developed a simpler model using a single attribute to define the \textit{Unary Constraint Model} (FCx-U), as represented in Equation \ref{eq:unarycon}. This allowed us to validate the functionality of the model. Furthermore, we created paired attribute combinations that could form logical constraints (constraints from pairs that directly influence each other with no exceptional rules), leading to the development of the \textit{Binary Constraint Model} (FCx-B), as illustrated in Equation \ref{eq:binarycon}.

We use the Adult dataset \cite{misc_adult_2}, which is thoroughly described in  Section \ref{sub:datasets} to form an illustrative example. This dataset includes attributes such as \textit{age} and \textit{education} which logically form the following constraint. If an individual attains a higher level of education, their age should also be greater. Based on this reasoning, we introduce Equations \ref{eq:unarycon} and \ref{eq:binarycon} into our feasibility evaluation model, enabling it to learn and preserve these constraints.

\begin{equation}
\label{eq:unarycon}
x_{\text{age}}^{cf} \geq x_{\text{age}}
\end{equation} 

\begin{equation}
\label{eq:binarycon}
\begin{array}{l}
\left(x_{ed}^{cf}>x_{ed} \Longrightarrow x_{\text{age}}^{cf}>x_{\text {age }}\right) \text { AND } \\
\left(x_{ed}^{cf}=x_{ed} \Longrightarrow x_{\text {age}}^{cf} \geq x_{\text {age}}\right)
\end{array}
\end{equation}

\subsection{Soft Causal constraints}
\label{softcausalconstraints}

Are we satisfied using only hard causal constraints to maintain feasibility? The answer is no. That's because we only look at specific pairs of features whose relationship if violated will automatically lead to infeasible CF examples. 

For this reason and after a series of experimentation on causal discovery techniques, we applied an intuitive method to introduce automatically extracted soft causal relationships, incorporating all possible pair-wise relations among every feature in a dataset. We employed PC-Algorithm \cite{Spirtes2000} before training, a causal discovery algorithm that met our criteria, to extract a Directed Acyclic Graph (DAG) from our data.  
In detail, the PC-Algorithm is a constrained-based method that extracts a DAG by testing conditional independencies between features. It first builds an undirected graph by removing edges based on statistical tests (skeleton), then orients the remaining ones to guarantee acyclicity and identify causal directions wherever possible. 

So, let $G$ be the causal graph and $A\in \{0,1\}^{d{\times}d}$ be derived from $G$, where $A_{p,q}=1$ indicates a directed causal influence from feature $Y_p$ to feature $Y_q$ and $A_{p,q}=0$ indicate that there is no directed causal influence from $Y_p$ to $Y_q$. All existing and non-existing feature relations in $A$, extracted from the graph $G$ will form the soft causal constraints for our model.

This procedure will be described in detail in the loss function section below.

\subsection{Sparsity}
\label{sec:sparsity}

An important aspect regarding the effectiveness of a CF explanation is \textit{sparsity}. When generating a CF example, we aim to modify the fewest possible features, to make the explanation more interpretable for the end-user and simultaneously guide the CF example to alter its class (see Figure \ref{fig:cfexampleandsparsity}) \cite{Sparsity}.  

Given a feature vector $\mathbf{x} = (x_1,x_2,...,x_d)$ and its corresponding CF example ${\mathbf{x}^{cf}} = (x_1^{cf},x_2^{cf},...,x_d^{cf})$, we define sparsity in  Equation \ref{eq:sparsity} as the total number of features that changed :
\begin{equation}
\label{eq:sparsity}
S\left(\mathbf{x}, \mathbf{x}^{cf}\right)=\sum_{p=1}^d 1\left(x_p \neq x_p^{cf}\right)
\end{equation}
where $1\left(x_p \neq x_p^{cf}\right)$ is an indicator function that equals to $1$ if $x_p$ is modified and $0$ otherwise. Our goal is to minimize $S\left(\mathbf{x}, \mathbf{x}^{cf}\right)$.

\section{Our Approach}
\label{methodology}

In this section, we provide an in-depth description of the proposed FCx framework, detailing the architecture of the VAE-based counterfactual generator, its interaction with the pretrained black-box classifier, the training algorithm, and the composite loss function used to guide the generation of feasible, valid, sparse, causally coherent and distributionally plausible counterfactual explanations.

\subsection{Architecture}
\label{sec:architecture}

\begin{figure}[htbp]
    \centering
    \includegraphics[width=0.45\textwidth]{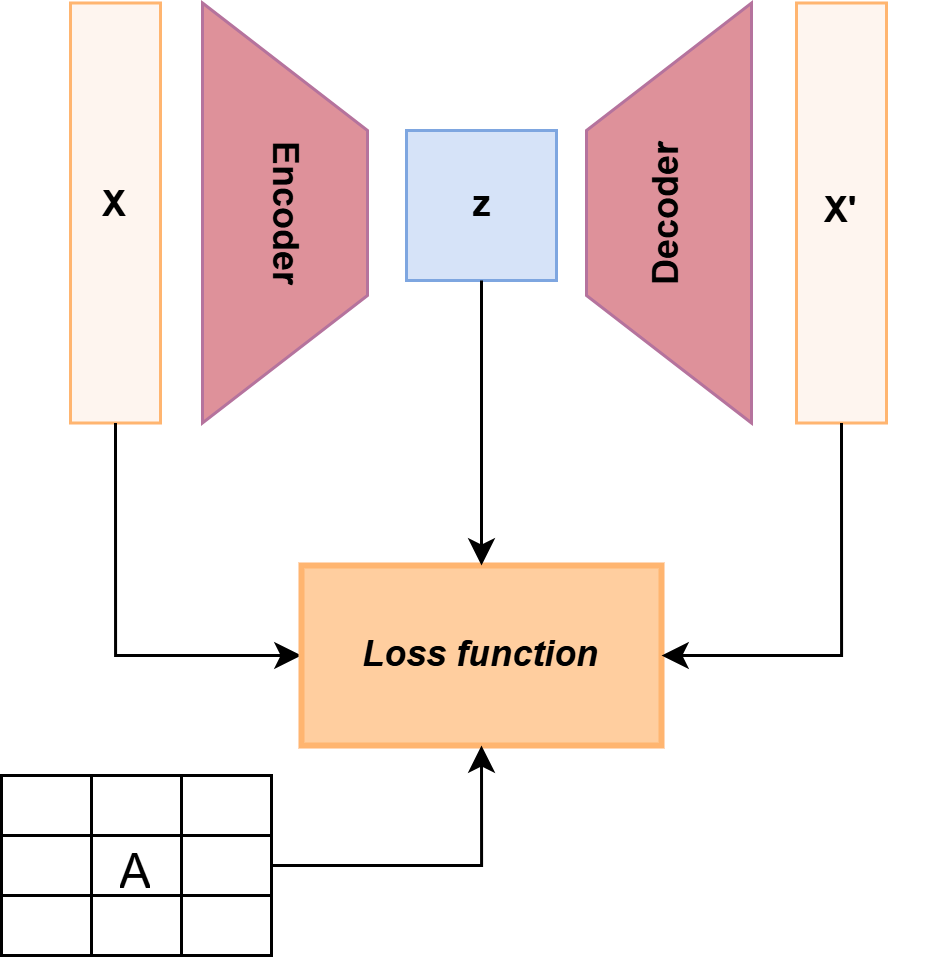}
    \caption{Overview of the FCx Architecture}
    \label{fig:architecture}
\end{figure}

In this section, we provide a high-level overview of our feasibility model architecture, as illustrated in Figure \ref{fig:architecture} as well as the algorithm of our training procedure. During training, input values $X$ are inserted into the encoder of a Variational Autoencoder (VAE) \cite{autoencoding}, which generates a lower-dimensional latent space representation. The decoder then attempts to reconstruct each input back to its original size, ensuring that every output serves as a counterfactual (CF) example. 

To guide the network in generating CFs examples that satisfy our main key criteria: proximity, validity, feasibility, sparsity, and density, a seven-part loss function is applied at each training epoch. The black-box model is used to predict the class of both the input and generated CF example, and it also contributes to the validity loss computation. A more detailed breakdown of the VAE’s architecture will be provided in  Section \ref{sub:implset}, along with Figure \ref{fig:detailedarchitecture}.

Initially, we train a black-box model, consisting of two linear layers, to classify input and output data points into two distinct classes. This gurantees compliance with the definition of a counterfactual (CF) example, where each input is paired with its desired (opposite) class. The output from this stage of the architecture is later incorporated into the validity loss function, serving as a pretrained model to accurately predict the correct class.

As a second and final step of our model, we utilize a Variational Autoencoder (VAE) to generate feasible CF examples. Our main goal is to create feasible CF examples from the input data points by satisfying both hard and soft causal constraints. For this reason, the VAE's encoder, which works as a bottleneck, is trained with heterogeneous data $X$ and produces a low- dimensional representation of them in latent space $z$. Finally, the output of $z$ must be reconstructed to its original dimensions in the decoder part of the VAE.  

Our model was trained with a multifactorial loss function, which will be described in detail in the next section.

\subsection{Loss function}
For the training of the VAE, we created a loss function that includes the main components from the problem definition defined in Section \ref{prob_definition}. Our loss function consists of seven parts in total, a KL- Divergence loss for the regularization of the latent space of the VAE, a proximal distance loss, a classification (validity) loss, a two part causal constraints loss, hard and soft, sparsity and LOF losses.  

Starting with the Equation \ref{cfdefinitionloss}  where the HingeLoss guarantees that the classifier $h$ assigns CF examples to the correct class with $\beta$ being a hyperparameter from a margin. The second term of this equation measures the proximity loss, which is a distance function $(L1/L2)$ that ensures that the generated CF example remains proximal to the original input. 
\begin{equation}
\label{cfdefinitionloss}
\operatorname{argmin}_{x^{cf}} \operatorname{HingeLoss}\left(h\left(x^{cf}\right), y^{\prime},\beta\right) + d(x,x^{cf})
\end{equation}
Since we are using a VAE (an encoder that maps the input features into a latent space and a decoder to reconstruct the CF example from the encoder's representation), we need a KL-Divergence loss to assure that the learned conditioned distribution remains close to a standard Gaussian prior $N(0,I)$, working as a regularizer of the model to ensure a smooth latent space, denoted as: 
\begin{equation}
    \operatorname{KL}\left(Q\left(z \mid x, y^{\prime}\right) \| P\left(z \mid y^{\prime}, x\right)\right)
\end{equation}. 
As for the hard causal constraints, we use $max(0,x-x^{cf})$ term for the unary constraint, where $x_p$ denotes the feature $p$ of input x and $max(0,(x_q^{xf}-x_q)-\alpha - \beta(x_p^{cf}-x_p) - min(0,\beta))$ for the binary one, where $\alpha,\beta$ are adjustable parameters and $x_q$ denotes the feature $q$ of input $x$. 

The soft causal constraints can be described as a two-fold Equation \ref{softconst} below: 

\begin{equation}
\label{softconst}
    \begin{aligned}
\mathcal{L}_{\text{soft}}(x^{cf})=\lambda_{\mathrm{nc}} \sum_{(p, q) \in \mathcal{E}^c}\left(\operatorname{Cov}\left(Y_p, Y_q\right)\right)^2+ \\
\lambda_{\mathrm{c}} \sum_{(p, q) \in \mathcal{E}}\left(\max \left(0, \theta-\operatorname{Cov}\left(Y_p, Y_q\right)\right)\right)^2
    \end{aligned}
\end{equation}

where $\mathcal{E}^c$ is a set of all the non-connected variable pairs $(p,q)$, $\mathcal{E}$ is the set of all connected ones and $\theta$ being a minimum covariance threshold that must be maintained from the connected variables and $Y$ is the adjacency matrix. 

The loss function $\mathcal{L}_{\text{sparsity}}= \left\|\mathbf{x}^{cf}-\mathbf{x}\right\|_0$, denotes sparsity between the input value and the generated CF example from their features, calculated with $L0/L1$ norm. 

The final part of the loss function is LOF which is calculated from the mean of the LOF scores for all points $n$ in the latent space $z$, with $k$ neighbors, as denoted in Equation \ref{lofloss} described in Section \ref{Feas_lof}. 

\begin{equation}
\mathcal{L}_{\text{LOF}}(z)=\frac{1}{n} \sum_{i=1}^n \operatorname{LOF}_k\left(z^{cfi}\right)
\label{lofloss}
\end{equation}

\subsection{Training algorithm}
\label{Training algorithm}

Our method is formalized in Algorithm~\ref{alg:fcex_abstract}, which trains a VAE-based generator for producing feasible counterfactual explanations. Given an input instance, the generator modifies only the mutable attributes while explicitly preserving immutable features. Counterfactual candidates are optimized through a composite objective that combines reconstruction-based proximity, predictive validity toward the target class, sparsity of feature changes, feasibility constraints, causal regularization, latent-space distributional consistency, and KL regularization. This encourages the generated counterfactuals to be actionable, causally coherent, close to the original instance, and consistent with the underlying data distribution.

\begin{algorithm}[H]
\caption{Training Feasible Counterfactual Explanations (FCx)}
\label{alg:fcex_abstract}
\begin{algorithmic}[1]

\State \textbf{Inputs:}
Dataset $\mathcal{D}$ (encoded); pretrained classifier $f$;
mutable indices $\mathcal{M}$, immutable indices $\mathcal{I}$;
causal adjacency matrix $A$; epochs $E$; batch size $B$;
weights $\lambda_{\mathrm{rec}}, \lambda_{\mathrm{val}},\lambda_{\mathrm{feas}},\lambda_{\mathrm{causal}},\lambda_{\mathrm{spar}}, \lambda_{\mathrm{lof}}$; margin $m$.

\State \textbf{Outputs:}
Trained generator parameters $(\phi,\theta)$.

\State \textbf{Initialize:} conditional VAE encoder $q_\phi(z\mid x_{\mathcal{M}},y)$ and decoder $p_\theta(x'_{\mathcal{M}}\mid z,y)$

\For{$e\gets 1$ \textbf{to} $E$}
  \ForAll{mini-batches $X\subset\mathcal{D}_{-}$ of size $B$}
    \State Target label $y \gets 1 - \operatorname*{arg\,max} f(X)$
    \State Sample latent $z \sim q_\phi(z\mid X_{\mathcal{M}},y)$
    \State Decode counterfactual candidates $x_{\mathcal{M}}' \sim p_\theta(\cdot\mid z,y)$

    \State Form full candidate $x'$ by enforcing immutables: $x'_{\mathcal{I}}\gets X_{\mathcal{I}}$
    \State $L_{\mathrm{rec}} \gets \textsc{Proximity}(x',X)$
    \State $L_{\mathrm{val}} \gets \textsc{ValidityHinge}(f(x'),y;m)$
    \State $L_{\mathrm{causal}} \gets \textsc{SoftCausalRegularization}(x'_{\mathcal{M}},A)$
    \State $L_{\mathrm{feas}} \gets \textsc{FeasibilityHardConstraints}(x',X)$
    \State $L_{\mathrm{spar}} \gets  \textsc{SparsityPenalty}(x',X)$
    \State $L_{\mathrm{KL}} \gets \textsc{KLDivergence}(q_\phi)$
    \State $L_{\mathrm{lof}} \gets \textsc{LOFRegularization}(z)$

    \State $L \gets \lambda_{\mathrm{rec}}\, L_{\mathrm{rec}} + L_{\mathrm{KL}}
        + \lambda_{\mathrm{val}}\, L_{\mathrm{val}}
        + \lambda_{\mathrm{feas}}\, L_{\mathrm{feas}} 
        + \lambda_{\mathrm{causal}}\, L_{\mathrm{causal}}
        + \lambda_{\mathrm{spar}}\, L_{\mathrm{spar}}
        + \lambda_{\mathrm{lof}}\, L_{\mathrm{lof}}$

    \State Update $(\phi,\theta)$ by backpropagation and an optimizer step
  \EndFor
  \State Optionally checkpoint and early-stop based on the loss
\EndFor
\State \Return $(\phi,\theta)$
\end{algorithmic}
\end{algorithm}

\subsection{Immutable Attributes}

A key aspect of training the VAE, which can significantly influence the quality of the generated counterfactual explanations, is the treatment of immutable attributes. Following \cite{REVISE}, we define an attribute as immutable when its value should remain unchanged in a counterfactual instance. For example, the attribute \textit{race} is considered immutable, since it cannot constitute an actionable change suggested by a counterfactual explanation. To ensure this constraint, immutable attributes are excluded from the generative modification process during VAE training. However, they are preserved from the original instance and reintroduced when constructing the final counterfactual candidate used for prediction. This guarantees that the generated explanations modify only actionable features while maintaining consistency with the individual's fixed characteristics.

\section{Experiments and results}

In this section, there will be an extended description of the datasets that we used, the preprocessing techniques that were applied to them. In  Section \ref{eval_metrics}, there will be a full presentation of all metrics that have been utilized to evaluate the model. In addition, an in depth overview of the experimental results of our method against a variety of different methods that generate CF examples will be presented for all four datasets. The source code and dataset are available
at \url{https://anonymous.4open.science/r/FCx_Finding-Feasible-Counterfactual-Explanation-3835}

\subsection{Datasets}
\label{sub:datasets}
We use four publicly available datasets as benchmarks, selected for their consistent use in previous studies. A summary of each dataset, including the total number of instances, the count of categorical, binary, and continuous attributes, the final number of inputs remaining after preprocessing, and the chosen decision/target class, will be provided. Lastly, we adopt an 80\%:10\%:10\% split for training, validation, and testing, respectively.

The Adult Income dataset \cite{misc_adult_2}, sourced from the UCI Machine Learning Repository, is a real-world dataset widely used for similar applications. It is designed to classify whether an individual’s income exceeds 50K per year. The dataset contains a total of 48,842 instances, of which we utilized 32,561. We selected 9 attributes, including 5 categorical, 2 binary, and 2 numerical/continuous features, resulting in a total of 9 features.

The KDD Census-Income dataset \cite{census_income_kdd} consists of weighted census data from the 1994 and 1995 Current Population Surveys conducted by the U.S. Census Bureau. Similar to the Adult dataset, it is used to classify whether an individual’s annual income exceeds 50K. The dataset includes 299,285 instances, of which we utilized 199,522. A total of 41 attributes were selected, comprising 32 categorical, 2 binary, and 7 numerical/continuous features, resulting in 164 features in total.

The Law School dataset \cite{Wightman1998LSACNL} contains law school admission records from 163 U.S. law schools in 1991. Its primary objective is to predict whether a candidate will pass the bar exam. The dataset originally consisted of 20,798 instances, of which we utilized 20,512. A total of 10 attributes were selected, including 1 categorical, 3 binary, and 6 numerical/continuous features, resulting in 21 features overall.

The Folktables database \cite{folktables} is sourced from the ASC PUMS. It can be used as a comparable replacement to the Adult dataset and it is designed to classify whether
an individual’s income exceeds 50K per year. The dataset
contains approximately 1.5M instances, of which we utilized
approximately 900K. We selected 9 attributes, including 5 categorical, 1
binary, and 2 numerical/continuous features, resulting in a total of 9 features.

\subsection{Implementation Settings}
\label{sub:implset}

This section provides a detailed description of the Variational Autoencoder (VAE) implementation, as illustrated in Figure \ref{fig:detailedarchitecture} below. Both the encoder and decoder consist of five linear layers, each followed by a ReLU activation function. To enhance training stability, batch normalization is applied after each linear layer. The fifth linear layer of the encoder estimates the mean ($\mu$) and variance ($\sigma^2$), with a sigmoid activation function applied to the variance to ensure its positivity. Additionally, a dropout rate of $10\%$ is introduced in each layer. The term \textit{Num. Features} refers to the size of each input processed by the encoder, including the target class. The latent space dimension $z$ is set to match the total number of features in each dataset, as discussed in Section \ref{sec:ablation}.

\begin{figure}[h]
    \centering
    \includegraphics[width=0.65\textwidth]{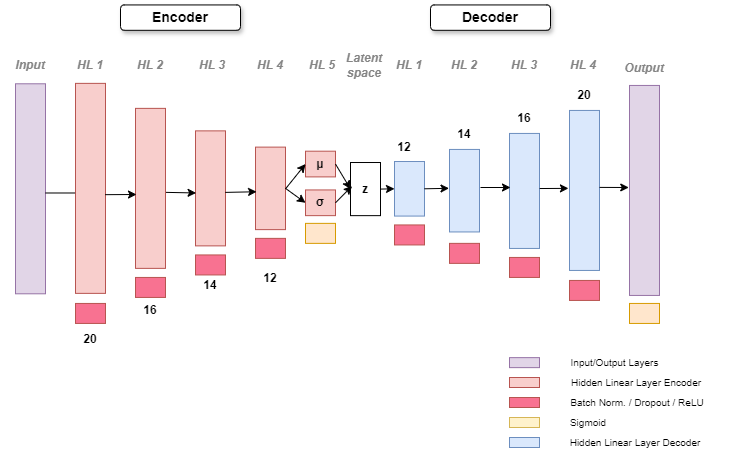}
    \caption{Our detailed VAE network}
    \label{fig:detailedarchitecture}
\end{figure}

\subsection{Preprocessing}\label{preprocess}
We applied the same techniques across all datasets during the preprocessing. First, any rows containing missing values were removed from the final dataset format. Continuous features were normalized to a range between 0 and 1, while categorical features were transformed using one-hot encoding. Binary attributes were converted to values of 0 and 1. The number of remaining instances has already been reported in  Section \ref{sub:datasets}. Finally, we extracted the causal graph $G$ from the preprocessed data, representing all connected and non-connected pairs in the adjacency matrix.

\subsection{Evaluation metrics}
\label{eval_metrics}
We used five metrics to evaluate the performance of our model regarding the feasibility of our generated CF examples. 

\begin{itemize}
\itemsep0em 
    \item \textbf{\textit{Validity:}} as “Validity", we measure the percentage (\%) of CF examples for which the predicted class aligns with the target class from the classifier. 
    \item \textbf{\textit{Feasibility Score:}} as “Feasibility score", we measure the percentage (\%) of the CF examples that satisfied the logical constraints, unary or binary.
    \item \textbf{\textit{Continuous and Categorical proximity:}} We define “Continuous proximity" as the average $L_1$ distance between the continuous features of the input value and the corresponding ones of the generated CF example and then we finally calculate the mean as $-\frac{1}{k} \sum_{i=1}^k \text { dist}\left(c_i, \boldsymbol{x}\right)$, where $(-1)$ multiplication, clarifies the proximity term, $k$ is the total number of CF examples and $x^{cfi}$ is a certain CF example. For "Categorical proximity", we first determine the total number of categorical feature alterations between the input value and its corresponding CF examples. The mean is then computed in the same manner as for continuous proximity.
    \item \textbf{\textit{Sparsity:}} We measure the number of features that changes for every input value to its corresponding CF example and from them, we calculate the mean “Sparsity score". 

    \item \textbf{\textit{Local Outlier Factor:}} With this technique, we determine how likely a data point is to be an outlier with respect to the k-closest instances in a multidimensional dataset. More specifically, we calculate the number of feasible CF examples that are further from the other CF points and the average LOF score which indicates to the density in the example space. 
\end{itemize}

\subsection{Comparative methods}
Both Tables \ref{table:ResultsDatasets_v1} and \ref{table:ResultsDatasets_v2} present the results of all methods applied to the four datasets. Mahajan Un/Bin \cite{Mahajan}, VC-Net \cite{VCNet}, FCEGAN \cite{FCEGAN} and CEILS \cite{CEILS} have been included as state-of-the-art methods to enable a direct comparison with our proposed approach. The methods REVISE \cite{REVISE}, C-CHVAE \cite{C-CHVAE}, CEM \cite{CEM}, and FACE \cite{FACE} were reproduced using the CARLA \cite{carla} Python library. CARLA is a benchmarking tool designed for counterfactual (CF) explanations, supporting various machine learning (ML) models and datasets. For the experiments involving DiCE \cite{DiCE}, we used one of its built-in models, named "random". DiCE is also a Python library that facilitates the generation of CF explanations to interpret ML-based predictions. A more detailed description of all these methods can be found in Section \ref{sub:related}.

\subsection{Results}
In this subsection, we present the final details regarding the datasets and provide a comprehensive overview of all experimental results, as summarized in both Tables \ref{table:ResultsDatasets_v1} and \ref{table:ResultsDatasets_v2}.

First, it is important to highlight that for the datasets, we used the \textit{age} attribute alone to define the unary constraint, and in combination with \textit{education}, it formed the binary constraint for the Adult, Folktables and KDD-Census datasets. In these cases, the \textit{Income} attribute served as the target or desired class. Similarly, for the Law School dataset, the unary constraint was based on the \textit{lsat} attribute, while the binary constraint was formed by combining \textit{lsat} with \textit{tier}. In this case, the \textit{Pass the bar} attribute was the target or desired class.

For the models from \cite{Mahajan} and our approach, we trained two separate models, one for the \textit{Unary constraint} and another for the \textit{Binary constraints}. In both cases, feasibility was used as both a learning parameter and an evaluation metric.

\begin{table*}[htbp]
    \centering
    \caption{Results on the Adult and Census datasets}

\begin{center}

\begin{tabular}{c|c|c|ccccc|cc}

\specialrule{.15em}{.15em}{.15em} 
& \multicolumn{1}{c|}{ \textbf{Methods} } & \multicolumn{1}{c|}{ \textbf{Validity}}& \begin{tabular}{c}\textbf{Cont.} \\ \textbf{prox.}\end{tabular}  & \begin{tabular}{c}\textbf{Categ.} \\ \textbf{prox.}\end{tabular} & \multicolumn{1}{c}{ \textbf{Sparsity} } & \multicolumn{1}{c}{ \textbf{LOF} } & \multicolumn{1}{c|}{ \textbf{Outliers} } & \begin{tabular}{c} \textbf{Feas.}\\ \textbf{Unary}
\end{tabular} & \begin{tabular}{c}\textbf{Feas.}\\ \textbf{Binary} \end{tabular} \\ \specialrule{.15em}{.15em}{.15em} 
\multirow{9}*{\rotatebox{90}{Adult}}
&  Mahajan Un.\cite{Mahajan}& 98.87 & -2.51 & -2.69 & 4.68 & 2.20& 54& 65.36 & - \\
  & Mahajan Bin.\cite{Mahajan}& 100.00 & -2.66 & -2.58 & 4.57 & 1.64& 39 & - & 70.38 \\
  & REVISE \cite{REVISE} & 54.07 & 
  -5.32 & -3.23 & 5.48 & 1.05& 0 &7.28 & 6.34 \\
  & C-CHVAE \cite{C-CHVAE}& 100.00 &  
  -3.41 &-4.25 & 6.25 & 16.14& 57 &48.01 & 49.03 \\
  & CEM \cite{CEM}& 74.00 &   
  -13.69 & -0.50 & 2.10 & 2.21&47 &21.88 & 21.80 \\
  & DiCE random \cite{DiCE} & 93.82 & 
  -3.83 & -0.64 &3.47 & 1.03& 0 &11.65 & 11.52 \\
  & FACE \cite{FACE} & 83.20 &   
  -4.82 & -3.29 & 5.52 & 2.34& 80 &32.12 & 30.82 \\

& VC-Net \cite{VCNet} & 100.00 & -4.44 & -2.69 & 4.48 & 6.64 & 46 & 26.36 & 23.54 \\
& FCEGAN \cite{FCEGAN} & 100.00 & -1.50 & -2.68 & 4.67 & 2.98 & 27 & 12.76 & 14.54 \\
& CEILS \cite{CEILS} & 28.49 & -13.44 & -1.83 & 3.00 & 9.53 & 65 & 52.34 & 55.37 \\

& \textbf{FCx-U} & 100.00 & -2.64 & -2.68 & 4.68 & 1.11 & 36 & \textbf{74.91} & - \\
& \textbf{FCx-B} & 100.00 & -2.67 & -2.57 & 4.55 & 1.23 & 30 & - & \textbf{82.92} \\ \specialrule{.15em}{.15em}{.15em} 

\multirow{9}*{\rotatebox{90}{Census}}
&  Mahajan Un.\cite{Mahajan}& 100.00 & -1.87 & -6.58 & 8.61 & 115.8 & 478 & 90.20 & - \\
&  Mahajan Bin.\cite{Mahajan} & 100.00 & -5.35 & -7.85 & 10.07 & 5.33 & 554 &- & 74.44 \\
&  REVISE \cite{REVISE} & 28.09 &   
  -7.39 & -11.27 & 20.30 & 116.09 & 188 &91.00 & 75.43 \\
&  C-CHVAE \cite{C-CHVAE}& 48.44 & 
  -7.66 &-11.94 & 19.74 & 1.12& 17&\textbf{98.87} & 76.61 \\
&  CEM \cite{CEM}& 86.68 &  
  -8.90 & -0.45 & 0.51 & 1.22& 250& 86.98 & 85.88  \\
&  DiCE random \cite{DiCE} & 97.51 &  
  -4.24 & -1.58 & 9.41 & 1.35 & 30 &93.50 & 88.36\\
&  FACE \cite{FACE} & 70.18 &    
  -8.11 & -8.30 & 12.38 & 1.71& 153&73.24 & 71.01 \\

& VC-Net \cite{VCNet} & 96.73 & -3.32 & -4.23 & 11.20 & 1.09 & 711 & 81.87 & 78.46 \\
& FCEGAN \cite{FCEGAN} & 100.00 & -9.73 & -7.66 & 15.32 & 6.59 & 696 & 75.06 & 74.34 \\
& CEILS \cite{CEILS} & 100.00 & -4.86 & -6.13 & 11.67 & 1.03 & 20 & 81.17 & 80.75 \\

& \textbf{FCx-U} & 100.00 & -1.56 & -8.12 & 10.50 & 1.92 & 396 & 93.96 & -\\
& \textbf{FCx-B} & 100.00 & -5.36 & -6.86 & 9.20 & 6.83 & 220 & - & \textbf{90.75}\\ \specialrule{.15em}{.15em}{.15em} 

\end{tabular}
\label{table:ResultsDatasets_v1}
\end{center}

    \label{table:ResultsDatasets_v1}
\end{table*}

\begin{table*}[htbp]
    \centering
    \caption{Results on Law-school and Folktables datasets}

\begin{center}
\begin{tabular}{c|c|c|ccccc|cc}

\specialrule{.15em}{.15em}{.15em} 
& \multicolumn{1}{c|}{ \textbf{Methods} } & \multicolumn{1}{c|}{ \textbf{Validity}}& \begin{tabular}{c}\textbf{Cont.} \\ \textbf{prox.}\end{tabular}  & \begin{tabular}{c}\textbf{Categ.} \\ \textbf{prox.}\end{tabular} & \multicolumn{1}{c}{ \textbf{Sparsity} } & \multicolumn{1}{c}{ \textbf{LOF} } & \multicolumn{1}{c|}{ \textbf{Outliers} } & \begin{tabular}{c} \textbf{Feas.}\\ \textbf{Unary}
\end{tabular} & \begin{tabular}{c}\textbf{Feas.}\\ \textbf{Binary} \end{tabular} \\ \specialrule{.15em}{.15em}{.15em} 
\multirow{9}*{\rotatebox{90}{Law-school}}
& Mahajan Un.\cite{Mahajan}& 100.00 & -12.75 & -1.13 & 7.13 & 1.00 & 0& 88.56 & - \\
&  Mahajan Bin.\cite{Mahajan} & 100.00 & -13.09 & -2.26 & 7.93 & 1.00 & 0 &- & 75.49 \\
&  REVISE \cite{REVISE} & 100.00 & 
  -19.55 & -2.23 & 7.81 & 1.71 & 6 & 71.87 & 69.51 \\
 & C-CHVAE \cite{C-CHVAE}& 100.00 &  
  -11.77 & -2.17 & 8.03 & 34.80& 17& 73.43 & 70.86 \\
&  CEM \cite{CEM}& 85.00 &   
  -5.01 & -1.52 & 2.68 & 1.04& 0& 56.38 & 55.25 \\
&  DiCE random \cite{DiCE} & 55.00 &   
  -4.68 & -1.16 & 5.64 & 1.01& 0& 85.48 & 24.24 \\
 & FACE \cite{FACE} & 100.00 &   
  -9.54 & -1.95 & 9.30 &1.04 & 0& 78.12 & 70.67 \\

& VC-Net \cite{VCNet} & 100.00 & -2.04 & -1.67 & 7.59 & 1.10 & 3 & 90.00 & 36.66 \\
& FCEGAN \cite{FCEGAN} & 83.33 & -11.19 & -1.54 & 7.51 & 1.02 & 0 & 92.66 & 47.05 \\
& CEILS \cite{CEILS} & 100.00 & -1.06 & -2.61 & 4.36 & 1.00 & 0 & 81.81 & 85.46 \\

& \textbf{FCx-U} &  100.00& -9.66 & -0.73 & 6.73 & 1.00 & 0 & \textbf{93.33} &  - \\
& \textbf{FCx-B} & 100.00& -13.48 & -1.54 & 7.00 & 1.00 & 0 &  - & \textbf{94.59 } \\

\specialrule{.15em}{.15em}{.15em} 

\multirow{9}*{\rotatebox{90}{Folktables}}
& Mahajan Un.\cite{Mahajan}& 100.00 & -1.75 & -2.56 & 4.54 & 1.07 & 102 & 81.33 & - \\
&  Mahajan Bin.\cite{Mahajan} & 100.00 & -1.05 & -2.97 & 4.82 & 1.00 & 75 &- & 67.95 \\
&  REVISE \cite{REVISE} & 100.00 & 
  -2.53& -2.92 & 4.81 & 1.00 & 0 & 37.12 & 40.26 \\
 & C-CHVAE \cite{C-CHVAE}& 100.00 &  
  -2.51 & -2.92 & 4.92 & 1.00 & 0 & 27.11 & 28.37 \\
&  CEM \cite{CEM}& 54.59 &   
  -2.23 & -0.55 & 0.55 & 1.67 & 294 & 54.95 & 57.56 \\
&  DiCE random \cite{DiCE} & 93.24  & -2.48  
   & -1.48 & 1.53 & 1.04 & 0 & 8.02 & 10.01 \\
 & FACE \cite{FACE} & 100.00 &   
  -4.11 & -3.42 & 5.20 & 1.06 & 0 & 4.07 & 4.43 \\

& VC-Net \cite{VCNet} & 100.00 & -4.70 & -2.37 & 4.34 & 1.03 & 420 & 34.85 & 35.06 \\
& FCEGAN \cite{FCEGAN} & 100.00 & -4.19 & -2.74 & 4.74 & 1.06 & 67 & 9.38 & 8.74 \\
& CEILS \cite{CEILS} & 24.40 & -1.52 & -1.64 & 2.75 & 1.02 & 0 & 59.01 & 61.52 \\

  & \textbf{FCx-U} &  100.00& -1.98 & -2.53 & 4.53 & 1.02 & 80 & \textbf{91.04} &  - \\
& \textbf{FCx-B} & 100.00& -1.15 & -2.99 & 4.98 & 1.09 & 64 &  - & \textbf{84.32} \\

\specialrule{.15em}{.15em}{.15em} 
\end{tabular}
\label{table:ResultsDatasets_v2}
\end{center}

    \label{table:ResultsDatasets_v2}
\end{table*}

Several conclusions can be drawn from the results presented in both Tables \ref{table:ResultsDatasets_v1} and \ref{table:ResultsDatasets_v2}. First, in almost all four datasets, our methods achieved the highest \textit{Feasibility} scores. Specifically, in the Adult dataset, FCx-U and FCx-B attained feasibility scores of $74.91\%$ and $82.92\%$, respectively. In the Law School dataset, they reached $93.33\%$ and $94.9\%$, while in Folktables we achieved $91.04\%$ and $84.32\%$ in FCx-U and FCx-B, respectively. In the KDD-Census dataset, FCx-B achieved a feasibility score of $90.75\%$; however, the C-CHVAE method outperformed our approach in the Unary model, achieving $98.87\%$. Despite this, the validity score for C-CHVAE was only $48.44\%$, indicating that our model with $100\%$ was better trained and more likely to generate valid counterfactual (CF) examples.

Regarding \textit{Validity}, our model consistently produced valid CF examples with the desired class, achieving a $100\%$ score for validity in every experiment. In terms of \textit{Sparsity}, the best results across all four datasets were obtained by CEM. However, these results are of lesser concern since high sparsity alone does not necessarily translate to better performance unless accompanied by strong validity and feasibility scores.

For LOF, almost all methods achieved values close to 1, suggesting that the generated CF examples closely resemble normal data points. Our methods demonstrated superiority in the Adult and  Law School datasets, achieving the best feasibility scores while maintaining competitive sparsity and LOF values. In contrast, the KDD-Census and Folktables datasets posed a greater challenge due to their combination of high-dimensional feature space and large sample size, resulting in elevated LOF values and more outliers.

Furthermore, our models consistently outperformed the method proposed by \cite{Mahajan}, which is most directly comparable to ours, in all evaluation metrics, while also surpassing more recent state-of-the-art methods such as \cite{FCEGAN} and \cite{CEILS}. Our methods, FCx-U and FCx-B can both be characterized as successful since they reduce sparsity to achieve high feasibility scores, they manage to create densely populated regions and eliminate outliers and the validity scores demonstrate that our model produces valid CF examples of the desired class.  

\subsection{Ablation Study}
\label{sec:ablation}
In this section, we assess the impact of the key components of FCx through a comprehensive ablation analysis that considers: (1) the dimensionality of the latent space $z$ for each dataset to determine the optimal setting for our model, (2) the ideal number of neighbors $k$ in LOF to ensure comprehensive coverage of all input instances in the latent space, (3) the robustness of the PC algorithm used to extract soft causal constraints, (4) an overview of the hyperparameter tuning for both our models (5) different combinations of our model’s components to assess their individual and collective contributions and (6) the behavior of all the components used in our model during training and validation.
 
\subsubsection{Dimensionality of latent space}

\begin{figure}[htbp]
    \centering
    \includegraphics[width=0.5\textwidth]{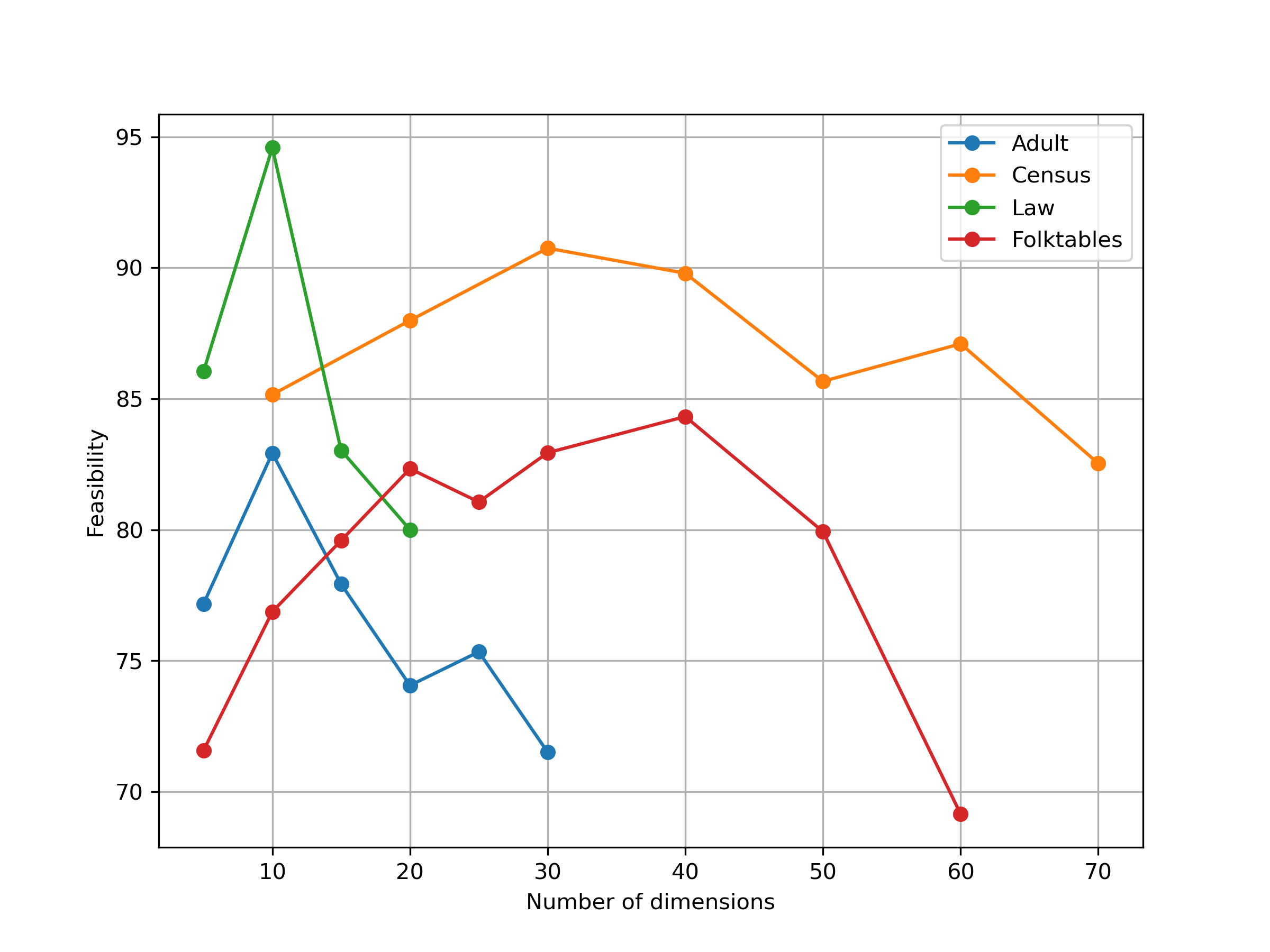}
    \caption{Experimentation of dimensions in latent space $z$}
    \label{fig:zdimentions}
\end{figure}

To determine the optimal number of dimensions in the latent space $z$, we experimented with sizes considering the total number of input features. As shown in Figure \ref{fig:zdimentions}, the best feasibility scores were obtained with 10 dimensions for Adult and Law, 30 for KDD-Census and 40 for Folktables given their larger size. All subsequent experiments maintained these latent space dimensions.

\subsubsection{Finding optimal number of neighbors for LOF}

\begin{figure}[htbp]
  \centering

  \begin{minipage}{0.45\linewidth}
    \centering
    \includegraphics[width=\linewidth]{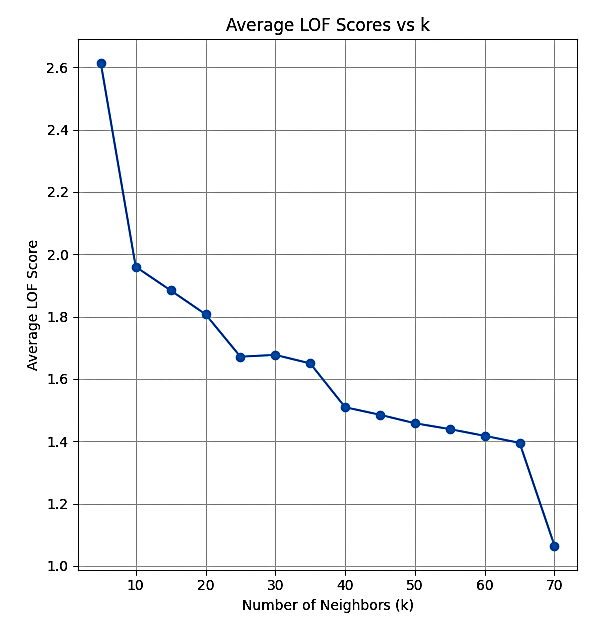}
    {\small (a) LOF scores\label{fig:k-values1}\par}
  \end{minipage}
  \hfill
  \begin{minipage}{0.45\linewidth}
    \centering
    \includegraphics[width=\linewidth]{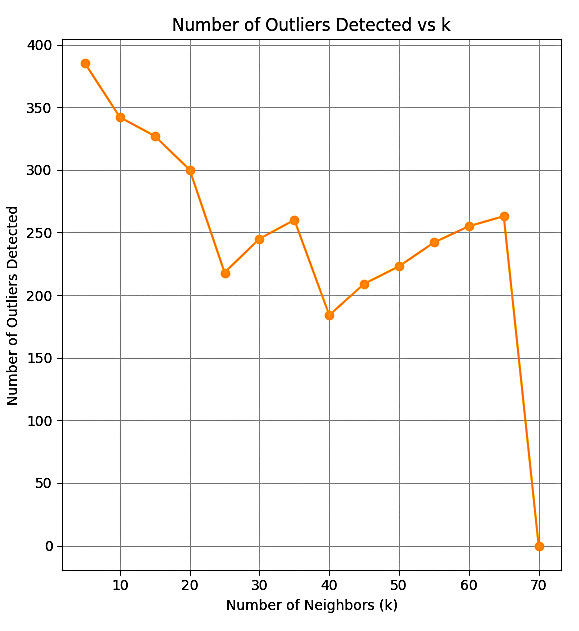}
    {\small (b) Number of outliers\label{fig:k-values2}\par}
  \end{minipage}

  \caption{Hyperparameter analysis of $k$ in LOF for the Adult dataset}
  \label{fig:k-values}
\end{figure}

For LOF, we experimented with various $k$ values, representing the number of nearest neighbors in KNN and LRD, to determine the most suitable for each dataset. As shown in Figure \ref{fig:k-values}, choosing an extremely small $k$ leads to excessive noise detection, while a very high $k$ reduces local outlier detection. To balance these effects, we set $k=20$ for the Adult dataset and followed a similar approach for KDD-Census ($k=50$), Law School ($k=15$) and Folktables ($k=70$), as mentioned in Table \ref{table:ImplDatasets}.

\subsubsection{Robustness analysis for PC algorithm}

Our framework is not dependent on the PC algorithm; it is compatible with adjacency matrices derived from any causal discovery method. We selected the PC algorithm due to its suitability for our data. However, its parameters, including the independence test and significance threshold, can be tuned specifically for each dataset, minimizing the risk of imposing unrealistic constraints from spurious edges. Empirical evaluations across four distinct datasets substantiate the reliability and strong performance of our method under diverse conditions.
To evaluate PC’s sensitivity in high-dimensional settings, we performed a robustness analysis by varying $\alpha$ (0.001, 0.01, 0.05, 0.1) and reporting mean and standard deviation for feasibility: 79.97\% ±2.71 (Adult), 91.31\% ±3.12 (Census), 91\% ±3.84 (Law) and 82\% ±2.63 (Folktables) with similarly low variance across other metrics, further confirming the method’s robustness.

\subsubsection{Hyperparameter tuning}
An overview of the hyperparameter tuning for each dataset employed in the FCx-U and FCx-B methods is provided in Table \ref{table:ImplDatasets}, including the chosen number of neighbors for LOF training.

\begin{table}[t]
\centering
\caption{Implementation settings}
\label{tab:implementation-settings}

\setlength{\tabcolsep}{4pt} 

\begin{tabular}{llccc}
\hline
\textbf{Dataset} &
\textbf{Method} &
\begin{tabular}{c}
\textbf{Learning} \\
\textbf{Rate}
\end{tabular} &
\textbf{Epochs} &
\textbf{$K$} \\
\hline Adult & FCx-U & $0.2$ & 25 & 20\\
  & FCx-B & $0.2$ & 50 & 20\\
\hline KDD & FCx-U & $0.1$ & 25& 50\\
Census & FCx-B & $0.1$ & 25 & 50\\
\hline Law & FCx-U & $0.2$ & 25 & 15\\
 School & FCx-B & $0.2$ & 50 & 15\\
\hline Folktables & FCx-U & $0.1$ & 25& 70\\
 & FCx-B & $0.1$ & 25 & 70\\
\hline
\end{tabular}
\label{table:ImplDatasets}
\end{table}

\subsubsection{Ablation study of loss components} 

We conducted an ablation study to assess the contribution of each component to our method and its impact on feasibility. We first trained a baseline model incorporating only feasibility constraints (proximity, validity, and hard causal constraints). Our findings in Table \ref{table:AblationStudy}, demonstrate that FCx, which integrates all proposed elements, outperforms all other configurations by achieving the highest feasibility score, highlighting the significance of each component. Regarding training time, incorporating additional components leads to increased computational cost, as expected, which is acceptable given our focus on improving feasibility rather than reducing training time.

\begin{table}[htbp]
\centering
\caption{Experimental study to measure the impact of the different components in our method on feasibility and the duration of the training process}
\label{tab:ablation}

\setlength{\tabcolsep}{4pt}

\begin{tabular}{llcc}
\hline
\textbf{Dataset} &
\textbf{Methods} &
\begin{tabular}{c}\textbf{Feas.}\\\textbf{Binary}\end{tabular} &
\begin{tabular}{c}\textbf{Duration}\\\textbf{(s/epoch)}\end{tabular} \\
\hline
\multirow{4}*{\rotatebox{90}{Adult}}
&  Baseline + sp & 77.54 & 3.21\\
  & Baseline + LOF & 77.64 & 41.06\\
  & Baseline + sp + LOF & 80.75 & 43.20\\
  & FCx-B& \textbf{82.92} & 43.44\\ \hline
  
\multirow{4}*{\rotatebox{90}{Census}}
&  Baseline + sp & 81.17  & 90.93\\
  & Baseline + LOF & 82.19 & 96.13\\
  & Baseline + sp + LOF & 88.63 & 100.44\\
  & FCx-B & \textbf{90.75} & 101.46\\ \hline

\multirow{4}*{\rotatebox{90}{Law-school}}
&  Baseline + sp & 89.19 & 0.47\\
  & Baseline + LOF & 90.54 & 3.35\\
  & Baseline + sp + LOF & 91.89  & 3.72\\
  & FCx-B & \textbf{94.59} &3.74\\ \hline

\multirow{4}*{\rotatebox{90}{Folktables}}
&  Baseline + sp & 65.20 & 291.1\\
  & Baseline + LOF & 61.88 & 303.3\\
  & Baseline + sp + LOF & 71.20 & 343.7  \\
  & FCx-B & \textbf{84.32} &386.4\\ \hline
\end{tabular}
\label{table:AblationStudy}
\end{table}

\subsubsection{Training and validation loss function and components}
\label{loss's diagram}

Finally, in the following figures, we report the evolution of all the components that compose the loss function of our model through time. The total training loss, comprising the ELBO and the additional regularization terms, is lower-bounded by design, which ensures that the optimization process remains stable throughout training. Although the non-convex nature of VAE-based models prevents guarantees of global optimality, standard gradient-based optimization methods can reliably converge to stationary points that correspond to local optima under reasonable smoothness assumptions. This behavior is also supported empirically by the reported curves. Overall, the model converges rapidly: the full loss decreases sharply during the first 5-10 epochs and then stabilizes, with training and validation curves remaining close throughout, indicating good generalization. The validity penalty collapses to near zero within the first few epochs, suggesting that the model quickly learns to satisfy the feasibility criteria for generated counterfactuals. Reconstruction and LOF losses decrease more gradually and stabilize after approximately 15-25 epochs, indicating continued improvement in reconstruction fidelity and adherence to the data distribution. Sparsity and soft causal loss terms improve more slowly, with a small train validation gap appearing in later epochs, where training continues to decrease while validation flattens, which is consistent with mild late-stage over-specialization. Finally, the KL term increases initially and then stabilizes, reflecting an increasing and then steady use of the latent representation as optimization proceeds.

\begin{figure*}[htbp]
\centering

\subfloat[Training and validation loss\label{fig:a}]{%
  \includegraphics[width=0.36\textwidth]{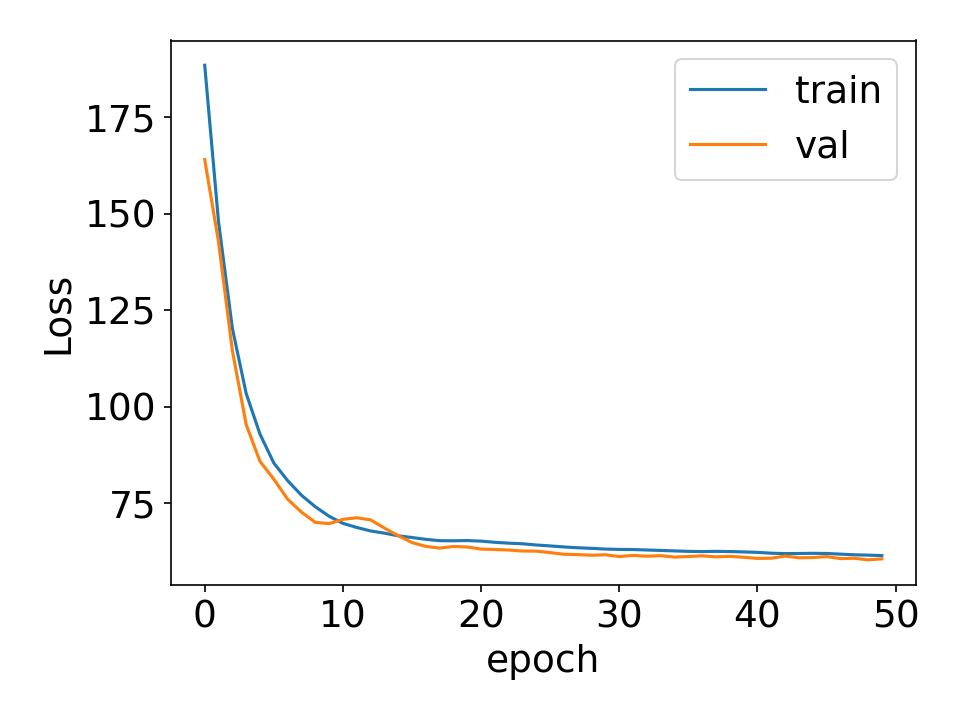}%
}\hfil
\subfloat[Hard constraint loss\label{fig:b}]{%
  \includegraphics[width=0.36\textwidth]{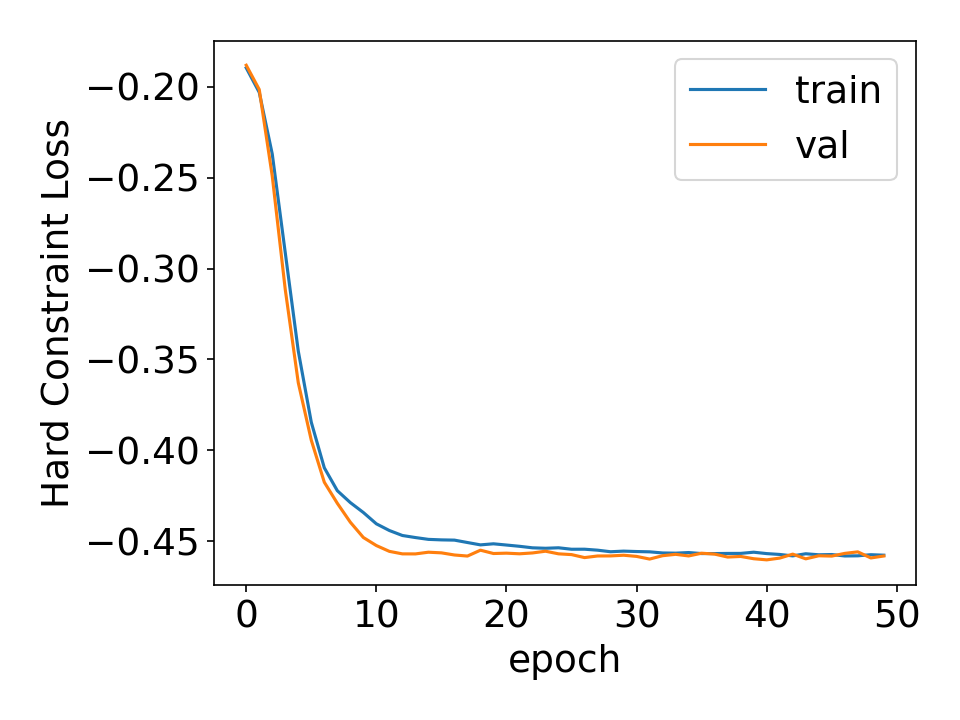}%
}\\[-0.9em]

\subfloat[KL component\label{fig:c}]{%
  \includegraphics[width=0.36\textwidth]{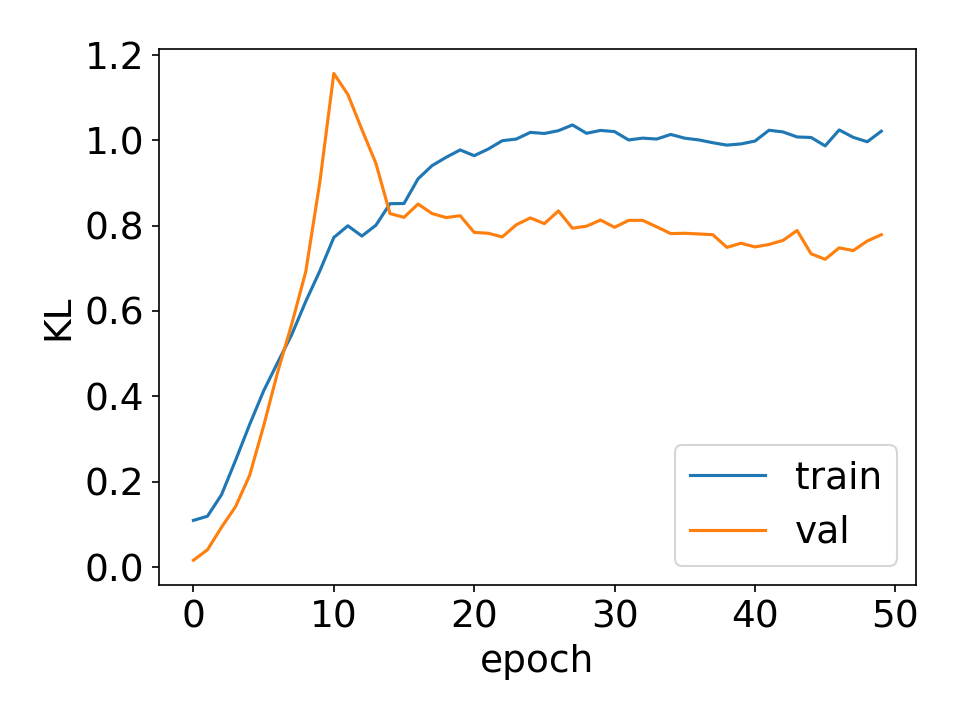}%
}\hfil
\subfloat[Reconstruction loss\label{fig:d}]{%
  \includegraphics[width=0.36\textwidth]{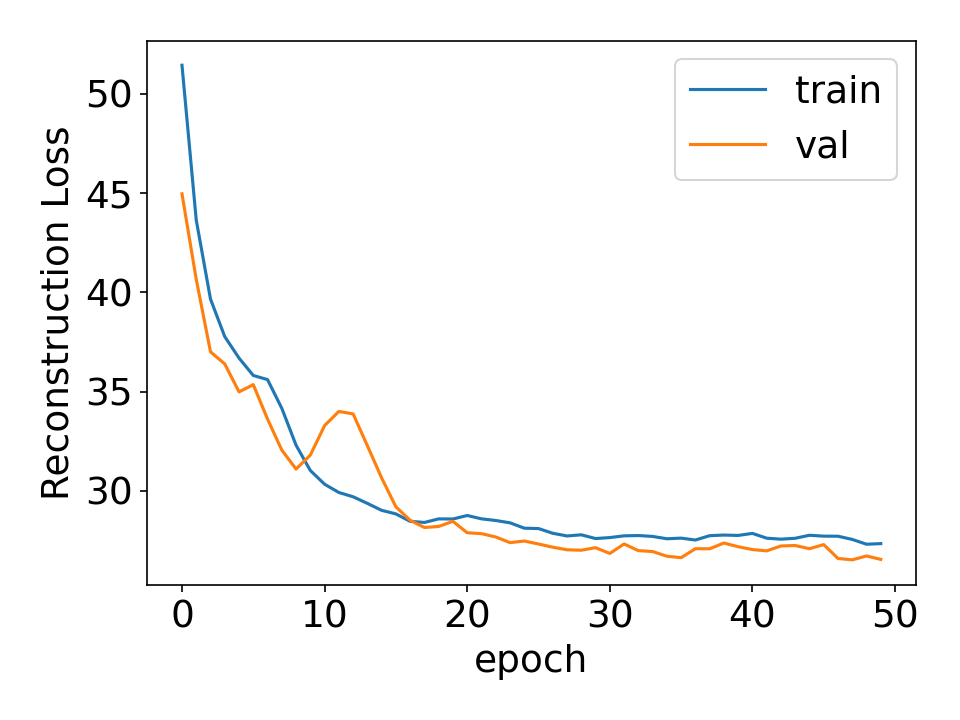}%
}\\[-0.9em]

\subfloat[LOF loss\label{fig:e}]{%
  \includegraphics[width=0.36\textwidth]{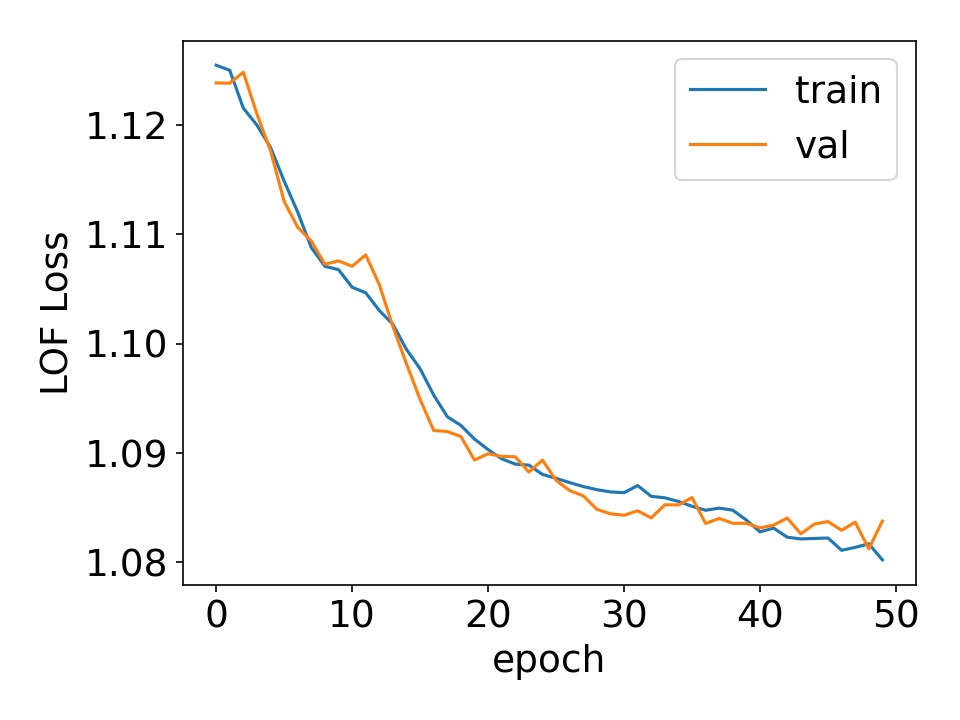}%
}\hfil
\subfloat[Soft causal loss\label{fig:f}]{%
  \includegraphics[width=0.36\textwidth]{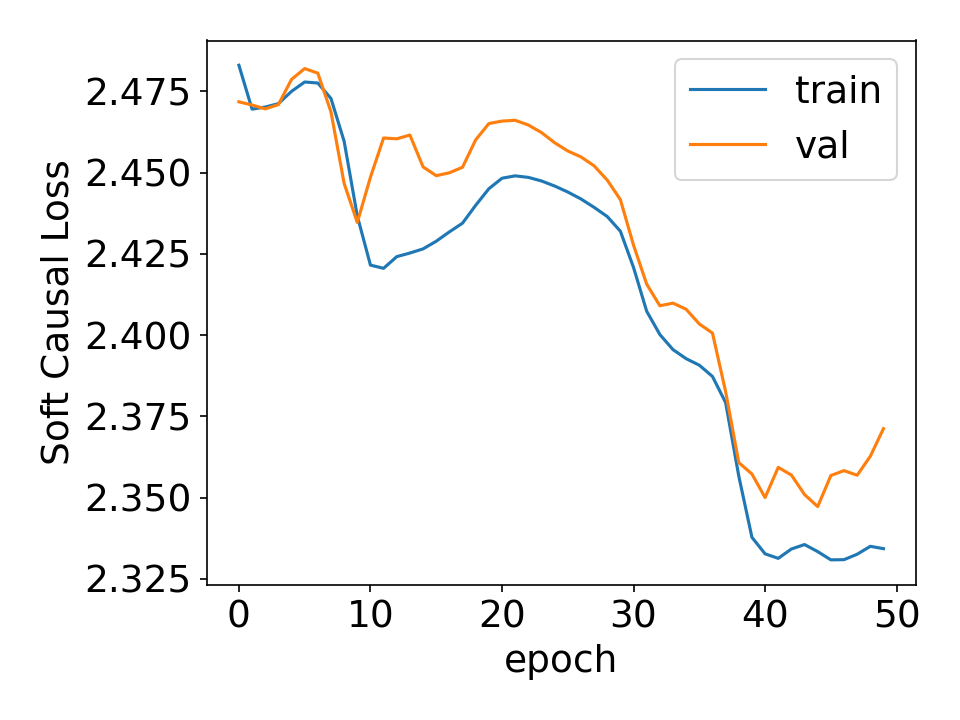}%
}\\[-0.9em]

\subfloat[Sparsity loss\label{fig:g}]{%
  \includegraphics[width=0.36\textwidth]{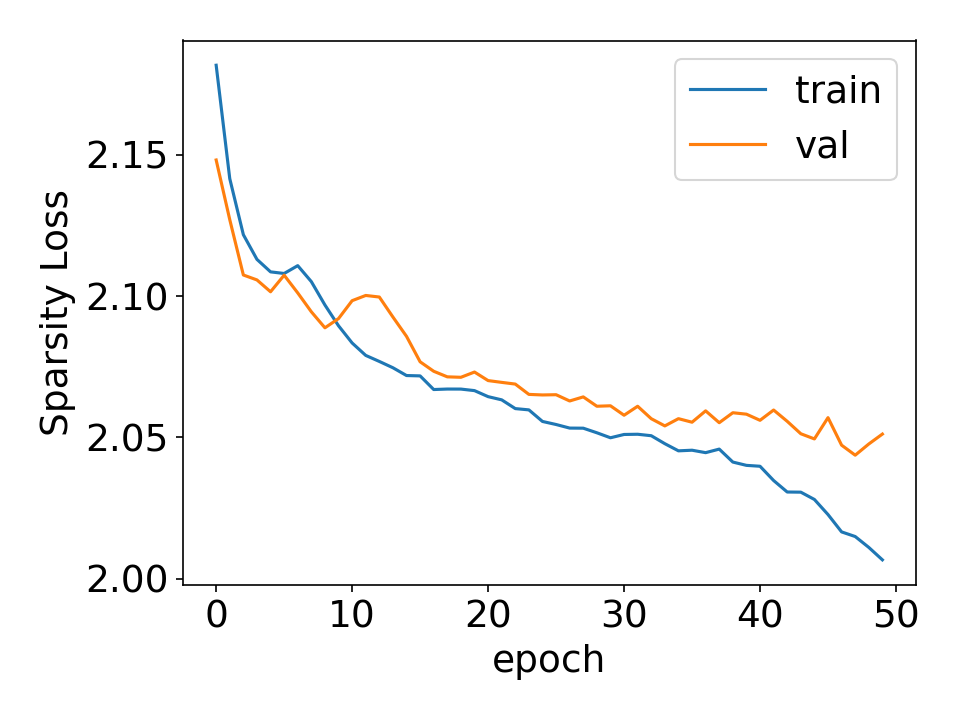}%
}\hfil
\subfloat[Validity loss\label{fig:h}]{%
  \includegraphics[width=0.36\textwidth]{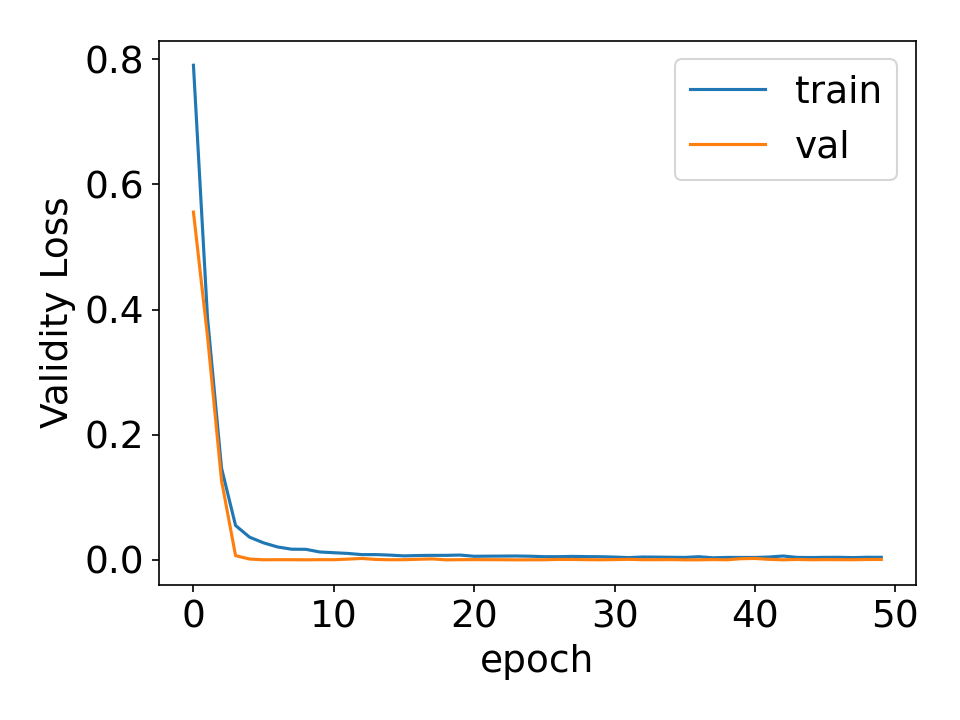}%
}

\caption{Training and validation losses and their components.}
\label{fig:big8}
\end{figure*}

\section{Conclusions}
Following an extensive series of experiments conducted across four diverse datasets, we derived several key insights regarding the performance and behavior of our proposed method. Firstly, our approach consistently outperformed existing baseline methods in terms of feasibility. This was achieved by incorporating both hard and soft causal constraints, enabling the model to effectively uncover and leverage previously unseen causal relationships among features. Such capability significantly reduces the need for manual intervention and underscores the method’s relevance and adaptability to real-world applications where domain knowledge may be limited or incomplete. Secondly, the integration of sparsity constraints into the counterfactual (CF) generation process yielded notable improvements. By promoting minimal but meaningful changes to feature values, sparsity ensured that the generated counterfactuals remained interpretable and actionable, aligning with realistic expectations of what can be changed in practice. Additionally, the application of the LOF technique further strengthened the reliability of our CF examples. By prioritizing generation within dense data regions, LOF helped prevent the creation of outliers or implausible instances, thereby enhancing the overall credibility of the counterfactuals. Lastly, our analysis maintained the critical importance of the underlying dataset structure in determining feasibility. The specific relationships and distributions of features within a dataset directly influence the causal constraints that guide CF generation. As a result, respecting and adapting to the structure of each dataset is essential to producing counterfactuals that are not only valid but also useful and constructive to the user.

\bibliographystyle{ACM-Reference-Format}
\bibliography{bibfile}

\end{document}